\documentclass[10pt,technote]{IEEEtran}
\usepackage{graphicx}
\usepackage{amsmath,amssymb} 
\usepackage{color}
\usepackage{mathptmx}
\usepackage{ragged2e}
\usepackage{url}
\usepackage{bm}

\ifCLASSOPTIONcompsoc
  \usepackage[nocompress]{cite}
\else
  \usepackage{cite}
\fi
\ifCLASSINFOpdf
\else
\fi
\newcommand{\Loss}{\mathcal{L}}
\newcommand{\Exit}{\mathcal{E}}

\begin{document}
%
\title{Measuring Human Perception to Improve Open Set Recognition}
%
%
%
%

\author{Jin~Huang,~\IEEEmembership{Student Member,~IEEE,} \\
Derek~Prijatelj,~\IEEEmembership{Student~Member,~IEEE} \\
Justin~Dulay,~\IEEEmembership{Student~Member,~IEEE} \\
and Walter~Scheirer,~\IEEEmembership{Senior~Member,~IEEE}

\IEEEcompsocitemizethanks{\IEEEcompsocthanksitem All of the authors are with the Department of Computer Science and Engineering, University of Notre Dame, Notre Dame, IN, 46556.\protect\\
Corresponding Author’s E-mail: walter.scheirer@nd.edu.}
}

%
%

\markboth{IEEE Transactions on Pattern Analysis and Machine Intelligence,~Vol.~X, No.~X, February~2023}%
{Shell \MakeLowercase{\textit{et al.}}: Bare Demo of IEEEtran.cls for Computer Society Journals}
%



\IEEEtitleabstractindextext{%

\begin{abstract}
The human ability to recognize when an object belongs or does not belong to a particular vision task outperforms all open set recognition algorithms. Human perception as measured by the methods and procedures of visual psychophysics from psychology provides an additional data stream for algorithms that need to manage novelty. For instance, measured reaction time from human subjects can offer insight as to whether a class sample is prone to be confused with a different class --- known or novel.
In this work, we designed and performed a large-scale behavioral experiment that collected over 200,000 human reaction time measurements associated with object recognition. The data collected indicated reaction time varies meaningfully across  objects at the sample-level.
We therefore designed a new psychophysical loss function that enforces consistency with human behavior in deep networks which exhibit variable reaction time for different images. As in biological vision, this approach allows us to achieve good open set recognition performance in regimes with limited labeled training data. 
Through experiments using data from ImageNet, significant improvement is observed when training Multi-Scale DenseNets with this new formulation: it  significantly improved top-1 validation accuracy by 6.02\%, top-1 test accuracy on known samples by 9.81\%, and top-1 test accuracy on unknown samples by 33.18\%.
We compared our method to 10 open set recognition methods from the literature, which were all outperformed on multiple metrics. 
\justifying
\end{abstract}

\begin{IEEEkeywords}
Computer Vision, Open Set Recognition, Novelty Detection, Visual Psychophysics,  Deep Learning
\end{IEEEkeywords}}

\maketitle

\IEEEdisplaynontitleabstractindextext

%
\IEEEpeerreviewmaketitle

\section{Introduction}
Open Set Recognition (OSR) is a task that is extremely challenging for computer vision algorithms~\cite{geng2020recent}, but something that can be effortlessly performed by humans without the need for very large labeled datasets~\cite{sunday2021novel}. In machine learning-based computer vision, OSR is defined as novelty detection coupled with closed set classification, where novelties are visual information not seen at training time that should not inhibit classification performance~\cite{walter_open_set}. At the class level, images from classes that are seen in both the training and testing phases are called known classes, while those that are not seen in training and only encountered in testing are considered to be unknown classes (\textit{i.e.}, novel classes). 

While some existing systems can detect unknown classes based on previous known knowledge~\cite{geng2020recent}, humans can not only recognize unknown instances but also avoid the under-generalization problem machine learning models face when fitting to known class data using just supervised labels~\cite{tenenbaum2011grow}. Psychologists study this phenomenon by using the methods and procedures of visual psychophysics to measure the human behavior associated with a particular task~\cite{lu2013visual}. Reaction time (RT) is one of the most diagnostic measurable behaviors because it reveals patterns of difficulty in the data (\textit{e.g.}, it is fast to recognize something that is familiar, while more ambiguous cases will take longer to be recognized).

\begin{figure}[t]
    \centering
    \includegraphics[width=\linewidth]{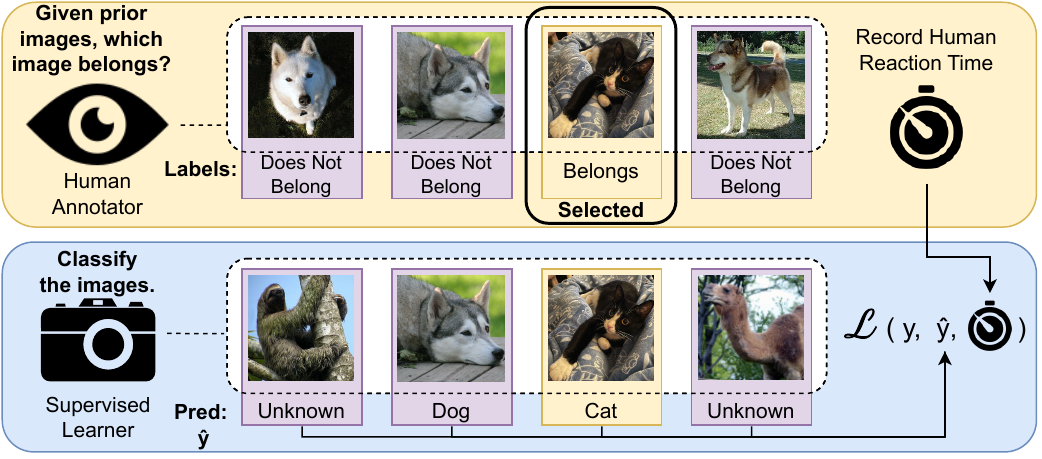}
    \caption{In this work, we conduct behavioral experiments to gauge the human perception of objects that belong or don't belong to a task. Through the use of visual psychophysics, human reaction time is measured across tasks (top panel). These measurements are then used in a novel psychophysical loss function to train biologically-inspired deep networks with a variable reaction time property, leading to better novelty detection and multi-class classification performance in open set recognition settings (bottom panel).
    }
    \label{fig:teaser}
    \vspace{-5mm}
\end{figure}

In this paper, we consider the possibility of incorporating measurements of human reaction time into the OSR process to improve model performance. The subtleties of human perception measured by psychophysics methods have already made in-roads in other computer vision tasks~\cite{Scheirer_2014_TPAMI,mccurrie2017predicting,grieggs2020measuring}, making this strategy an attractive target for OSR.
The fact that humans can separate objects that belong to a task from objects that do not in a better way than machines indicates that incorporating psychophysical measurements of human reaction time into the model training process may provide more information to learn from, and thus better performance.
The challenge is in designing a biologically-inspired neural network architecture with its own variable reaction time and conditioning it to match human behavior as closely as possible. 

The first step to solving this problem is collecting data that are useful in the context of novelty management.
Boult et al.~\cite{boult2020unifying} have recently argued that much of the existing OSR work consists of ill-defined novelty problems, making both data collection and algorithm design more difficult than is necessary. Thus they introduce a formal taxonomy of novelty and suggest that the perception of novelty can be decoupled from the ``ground-truth'' novelty value of objects in the environment. This means that something can be novel to an agent (artificial or human) that may not be truly novel in the environment (\textit{i.e.}, it has not been learned yet). In-line with this, we designed a human behavioral experiment which gauged the human perception of class-level information via reaction time measurement (Fig.~\ref{fig:teaser}, top panel). We collected over 200,000 psychophysical reaction time measurements using this experiment design for a partition of ImageNet containing 335 total classes~\cite{UMD_dataset}. Compared to prior work in OSR~\cite{Bendale_2016_CVPR,Rudd_2018,zhou2021learning,crosr,cacosr}, this dataset is far more realistic and challenging than the open set partitions of MNIST, SVNH, CIFAR10 and other small datasets that are commonly found in the literature. But it is intentionally more limited in available training images compared to the common ImageNet-based OSR training regimes.

The second step is the formulation of the machine learning strategy. After obtaining knowledge of the correlation between reaction time and OSR performance for individual images, we designed a new psychophysical loss function that can enforce more consistent behavior between humans and machines when training a deep neural network, thus improving the performance on OSR problems (Fig.~\ref{fig:teaser}, bottom panel).
To ensure the best match between the newly introduced loss function and the task, we chose to use the Multi-Scale DenseNet~\cite{huang2018multiscale} (MSD-Net) architecture as a proof-of-concept.
MSD-Net has five classifiers at different depths in its architecture, making it possible to measure model reaction time by observing the classifier that serves as the exit point for a particular image.
We believe that coupling large-scale psychophysically annotated data with a loss function that makes use of behavioral measurements like reaction time to induce similar behavior in models is a viable path forward for the OSR problem. 


\textbf{In summary, the major contributions of this paper are:} \textbf{(1)} 
A new dataset for the OSR problem containing reaction time measurements for a challenging partition of ImageNet.
This data is analyzed to draw out patterns of difficulty not evident from the original class-level labels.
This is the first large-scale human behavior study carried out to explore the use of psychophysical measurements for OSR, and it provides a wealth of data for other researchers working on this problem and related ones\footnote{All data and code will be released after publication.}.
\textbf{(2)} A general strategy for utilizing measured human reaction time for training deep networks for OSR, implemented as a psychophysical loss function that can improve the performance of any neural network architecture that supports variable reaction times for different inputs.
\textbf{(3)} A specific implementation of the loss strategy that takes advantage of the special structure of the MSD-Net architecture.
\textbf{(4)} Extensive experimentation over the OSR partition of ImageNet that balances class diversity and limited training data, with comparison made to 10 recent OSR approaches from the literature.
MSD-Net models that are trained with the proposed psychophysical loss function are shown to significantly outperform prior work. 
\section{Related Work}

\textbf{Open Set Recognition.} The development of capabilities that can detect that a sample is novel is a classic problem within the broader field of pattern recognition~\cite{ruff2021unifying}. But merely detecting novelty is a limited task --- classification is far more useful mode of operation in machine learning. Most recognition work in computer vision has been conducted in a closed set classification mode, meaning all classes are known at both training and testing time. In contrast, OSR is a more realistic scenario, where partial  knowledge of the world is present during training, and unknown phenomena are guaranteed to appear during testing. The difficulty has been in finding effective approaches for distinguishing between known and unknown samples~\cite{geng2020recent}.

Scheirer et al. initially formalized the OSR problem~\cite{walter_open_set} and introduced the concept of~\textit{openness}, which characterizes the potential difficulty of open set problems. Importantly, they defined a notion of open space risk, based on the assumption that the farther away a sample is from the support of known training data in a feature space, the riskier it is to assign a known class label to it. Following this definition, a series of standalone classifiers was developed, including the 1-vs-Set Machine~\cite{walter_open_set}, $P_I$-SVM~\cite{pi-svm}, and Weibull-Calibrated SVM (W-SVM)~\cite{wsvm}. Moving beyond SVM-like classifers, Rudd et al. proposed the Extreme Value Machine (EVM)~\cite{Rudd_2018}, which is an OSR classifier based on the statistical Extreme Value Theory (EVT) that supports incremental learning.

OpenMax~\cite{Bendale_2016_CVPR} was the first OSR classifier that could be trained with a deep neural network, adapting EVT concepts to the activation patterns in the penultimate layer to perform OSR. Extending OpenMax, Generative OpenMax~\cite{g_openmax} trains a deep network with synthesized unknown data. Similarly, Counterfactual Images for
Open Set Recognition (OSRCI)~\cite{Neal_2018_ECCV} uses a Generative Adversarial Network (GAN) to generate unknown images that are close to the training images but technically do not belong to any training class.
Class Reconstruction Learning for Open Set Recognition (CROSR)~\cite{crosr} utilizes latent representations for reconstruction to improve performance on unknown samples.
The Class Conditioned Auto-Encoder for Open Set Recognition (C2AE)~\cite{c2ae} divides the training procedure into two sub-tasks, closed set classification and open set identification, and uses EVT to find a threshold for filtering unknown samples.
Class Anchor Clustering for Open Set Recognition (CAC-OSR)~\cite{cacosr} uses a distance-based loss function that forces known classes to form tight clusters around anchored class-dependent centers.
Multi-Task OSR~\cite{multi-task-osr} combines a classifier and a decoder network with a shared feature extractor network within a multi-task learning framework for OSR.

Outlier Exposure (OE) is another strategy for OSR, where outlier data is provided to the model during the training process so it can generalize and ideally detect unknown samples. It is possible to leverage outlier samples to improve anomaly detection by training anomaly detectors against an auxiliary dataset of outliers~\cite{Deep_Anomaly_Detection_with_Outlier_Exposure}. Dhamija et al.~\cite{Reducing_Network_Agnostophobia} introduced a new evaluation metric that focuses on comparing the performance of multiple approaches in scenarios where unknowns are presented and proposed novel losses that are designed to maximize entropy for unknown inputs. Kong and Ramanan~\cite{openGAN}  introduced OpenGAN, which augments training outliers with fake unknown data synthesized by a generator trained to fool the discriminator.

Considering all of the above approaches, the human capacity for managing novelty has been missing. The previous work has focused on formalizing the OSR problem and developing  methods that can improve OSR performance. However, none are informed by the measurement of human perception. This is surprising, considering that humans have a much stronger ability to perceive what belongs or doesn't belong to a task than computer algorithms. Different from previous OSR work, we suggest the use of visual psychophysics as a path forward. 

\textbf{Visual Psychophysics for Computer Vision.} Some researchers have attempted to draw a connection between the human visual system and artificial neural networks, looking into the differences between various visual systems~\cite{sensitivity}, as well as comparing the consistency between human vision and deep networks~\cite{DBLP:journals/corr/EberhardtCS16}.  Furthermore, a growing amount of work has been studying how human behavior and perception can be related to solving computer vision problems. For example, DiCarlo et al.~\cite{DICARLO2012415} highlighted how the human brain can perform object recognition despite a difficult environment and many variances in the setting. After reviewing the evidence from human behavior to neural recordings, they proposed that neuronal and psychophysical data is necessary for better computational models. Medathati et al.~\cite{MEDATHATI20161} presented an overview of computational approaches to biological vision, and showed how new computer vision methods can be developed from biological insights. And studies have found that the human visual system takes different amounts of time to respond to stimuli of varied difficulty~\cite{A_Computational_Perspective, Duncan1989VisualSA}.
All of these works touched on the notion that psychophysical methods show promise for computer vision algorithm development, but from the perspective of biology. 

From the perspective of computer vision, Scheirer et al. proposed the idea of \textit{Perceptual Annotation}~\cite{Scheirer_2014_TPAMI} to make use of psychophysical measurements collected via crowdsourced means to train more accurate SVM classifiers. McCurrie et al.~\cite{mccurrie2017predicting} collected behavioral data reflecting human judgments of subjective facial attributes in order to model them. Zhang et al.~\cite{zhang2018agil} suggested the
use of human gaze measurements, another  behavioral measurement type, for improving performance in  object-related tasks. RichardWebster et al.~\cite{brandon_psyphy,RichardWebster_2018_ECCV} proposed an evaluation framework for visual recognition models that constructs item-response curves made up of individual stimulus responses to find perceptual thresholds. 

Most closely related to the work in this paper is that of Grieggs et al. \cite{grieggs2020measuring}, which introduces a psychophysical loss formulation for training artificial neural networks. In that work, behavioral experiments were conducted to collect reaction time data associated with the ability of reading in order to improve handwritten character recognition in historical documents. There are a few major differences between our work and~\cite{grieggs2020measuring}:
\textbf{(1)} We  develop a psychophysical loss function that incorporates several data streams and error measures.
\textbf{(2)} Our work focuses on the general area of image classification instead of the relatively niche area of historical document processing.
Not much expertise is required for our task, thus it is possible for us to conduct larger-scale human data collections.
\textbf{(3)} Our goal is to utilize the psychophysical loss to improve a deep network's performance on OSR, instead of just achieving better closed set classification performance. 
\section{Psychophysical Study Of Known Data}
\label{sec:study}

To better understand how humans perceive familiar images in order to be tolerant to novelty, we designed and conducted a study using Amazon Mechanical Turk to collect human reaction time (RT) data\footnote{Approved by University of Notre Dame IRB under protocol 18-01-4341.}. As the source image data, we used a subset of ImageNet~\cite{ILSVRC15} that contains 335 classes (called ImageNet335 in this paper)~\cite{UMD_dataset}, which was created by the University of Maryland, Carnegie Melon University, and Columbia University as part of the DARPA Science of Artificial Intelligence and Learning for Open-world Novelty (SAIL-ON) program. For the purpose of novelty detection, we partitioned that dataset into two categories for machine learning. \textbf{(1)} \textit{Known Classes}: the classes with distinctly labeled positive training examples that appear at both training and testing time; these are the non-novel classes. \textbf{(2)} \textit{Unknown Classes}: the classes that are not labeled and are unseen in training; these are the novel classes used in testing.

With respect to the specific breakdown, we randomly selected 42 classes as unknown classes from ImageNet335, and the remaining 293 classes were used as known classes. In this study, we collected reaction time measurements for the detection of specific known samples in the midst of images from other classes. Knowing from previous work that a mix of original and psychophysically annotated training data tends to yield good results~\cite{Scheirer_2014_TPAMI, grieggs2020measuring}, we randomly chose 40 classes from the known classes to use in the human behavior experiments.

%
%


\textbf{Study Design.} While \emph{novelty} can be defined in different ways depending on the task and context, the problem we are trying to address in this research is class-level novelty. We designed a study that is suitable for collecting human reaction time measurements associated with known classes; we do this because unknown classes should not appear at training time in any capacity for a fair evaluation. Accordingly, we need a task that can provide information about the patterns of difficulty associated with known training samples as they interact with other classes. Here we can use other known class data as stand-in material for novel samples as RT measurements, which reflect difficulty, are made. 

\begin{figure}
    \centering
    \includegraphics[width=\columnwidth]{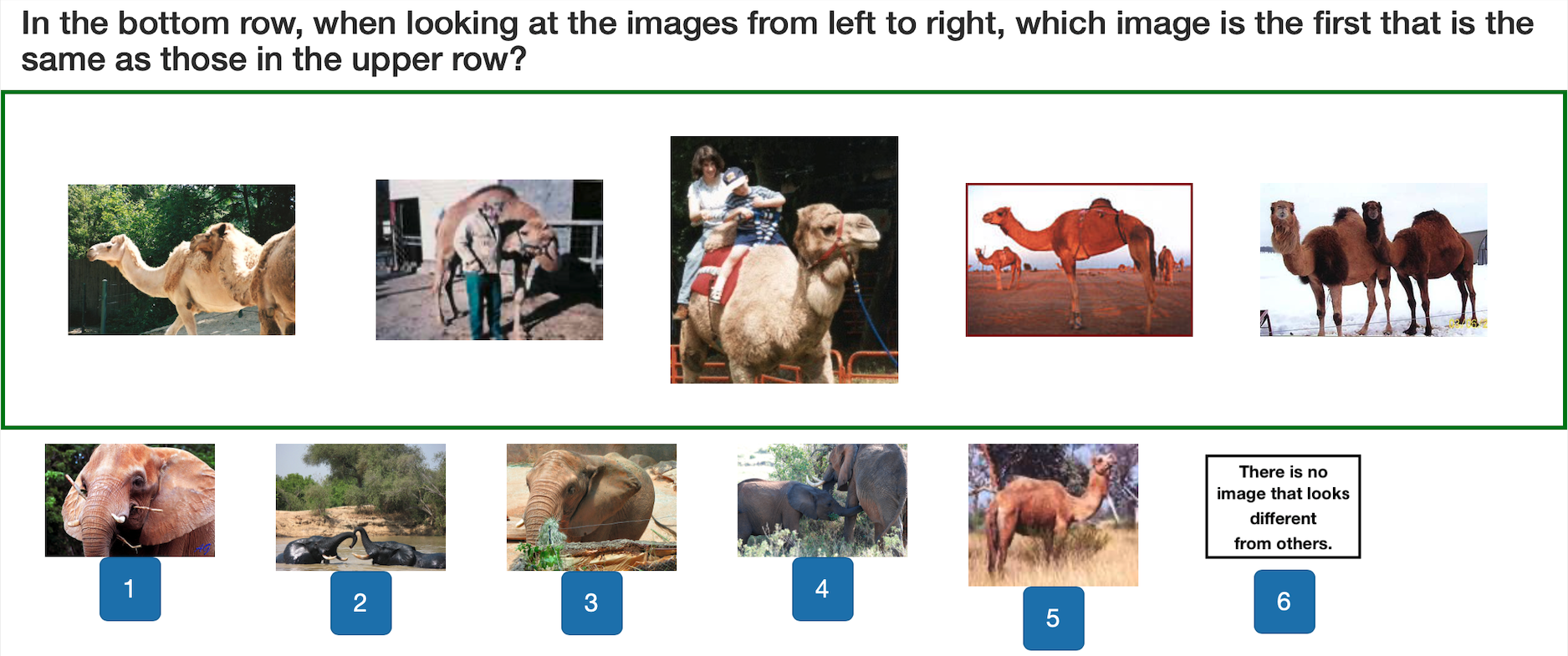}
    \caption{
        A sample of one of the survey questions for collecting human reaction time for known images.
        The subjects are asked to look at both rows of images and select the first image in the bottom row that is of the same object class as the reference class in the top row. It is possible that no image from the reference class is shown, hence the sixth option in the bottom row. In this specific example, the subjects should select the fifth image.
    }
    \label{fig:data_survey_design}
    \vspace{-5mm}
\end{figure}


Each Mechanical Turk survey contained 25 questions, and subjects received 25 cents for completing a full survey.  Fig.~\ref{fig:data_survey_design} shows an example of one survey question (the full survey process is detailed in Supp.~Mat.~Sec.~1.1). In each question, two rows of images are shown to a subject. In the top row, five images are shown within a green box and are treated as reference images --- all of them from the same class. In the bottom row, five additional images are shown. The task is for the subject to look at the images in the bottom row sequentially from left to right in order to find the first image that belongs to the reference image class. If the subject believes that such an instance isn't present in the bottom row, then they can indicate that as their response. A timer is started when a question populates the page and is stopped when the subject makes a decision. This is the recorded reaction time that is associated with the image that belongs to the reference class in the question. If the correct answer is the sixth option (a quality control measure), then the subject's recorded answer and RT will not be used for machine learning training.

We consider the reference images in the first row of a question as training data for a human, and the images in the second row as testing data. Training images for our machine learning task are drawn from the correct known class images from the second rows across questions and surveys. The intuition behind this design is that RT on known samples provides more information on what is non-novel beyond a class label, and if a model has a better understanding of what is known, it should at the same time be able to better judge what could be unknown by increasing the separation between these two categories. This is consistent with a recent observation that novelty detection in deep networks works by detecting the absence of familiar features as opposed to the presence of novelty~\cite{dietterich2022familiarity}. Taking this idea a step further, the data from our study can be utilized effectively given a deep network with a notion of variable reaction time for each input~\cite{huang2018multiscale}, which is trained to enforce behavioral fidelity with humans.



We pick one known class for the first row training data and another known class as a stand-in novel class that is different from the training data for the second row testing data. To be specific, in a question whose answer is not the sixth option, there are 5 images that are from a single class in the first row; in the second row, there are 4 images that are from a different class and 1 image that is from the same class as the first row. Such a design is straightforward for subjects to understand because there are only 2 classes shown at a time in a question, thus we expect to minimize noise in the results.
Assuming we pick $n$ classes for the surveys, each class is paired with itself and the rest of the $n-1$ classes. Using each pair, 20 questions are created, and based on the above there are $n^2$ different sets of surveys for the $n$ classes chosen. In the experiments conducted there are 40 known classes, thus there are 1,600 sets of different surveys in total.

To filter out bad submissions, we also include five control questions in every survey, which means each survey has 25 questions in total. These control questions are designed to be trivially easy so that any diligent worker will most likely get them correct. We allow the subjects to answer at most two control questions incorrectly. Answers provided by unreliable subjects are filtered out and are not considered for the machine learning experiments. Subject were not able to answer the same question twice. 
 Valid response times had to be under 28 seconds (the maximum RT after removing the largest RT values reflecting 5\% of the total data) on the assumption that anything longer reflected an inattentive subject. Although we instructed the subjects to look at the images from left to right sequentially, there was no control to enforce this on their side. Thus we require that 5 subjects complete each question so that noise can be smoothed out by considering average reaction times for each image. To ensure the order of the images does not affect the reaction time entries, all the images are selected randomly when the survey is generated and the images do not follow any pre-set order.



\begin{figure}
    \centering
    \includegraphics[width=\linewidth]{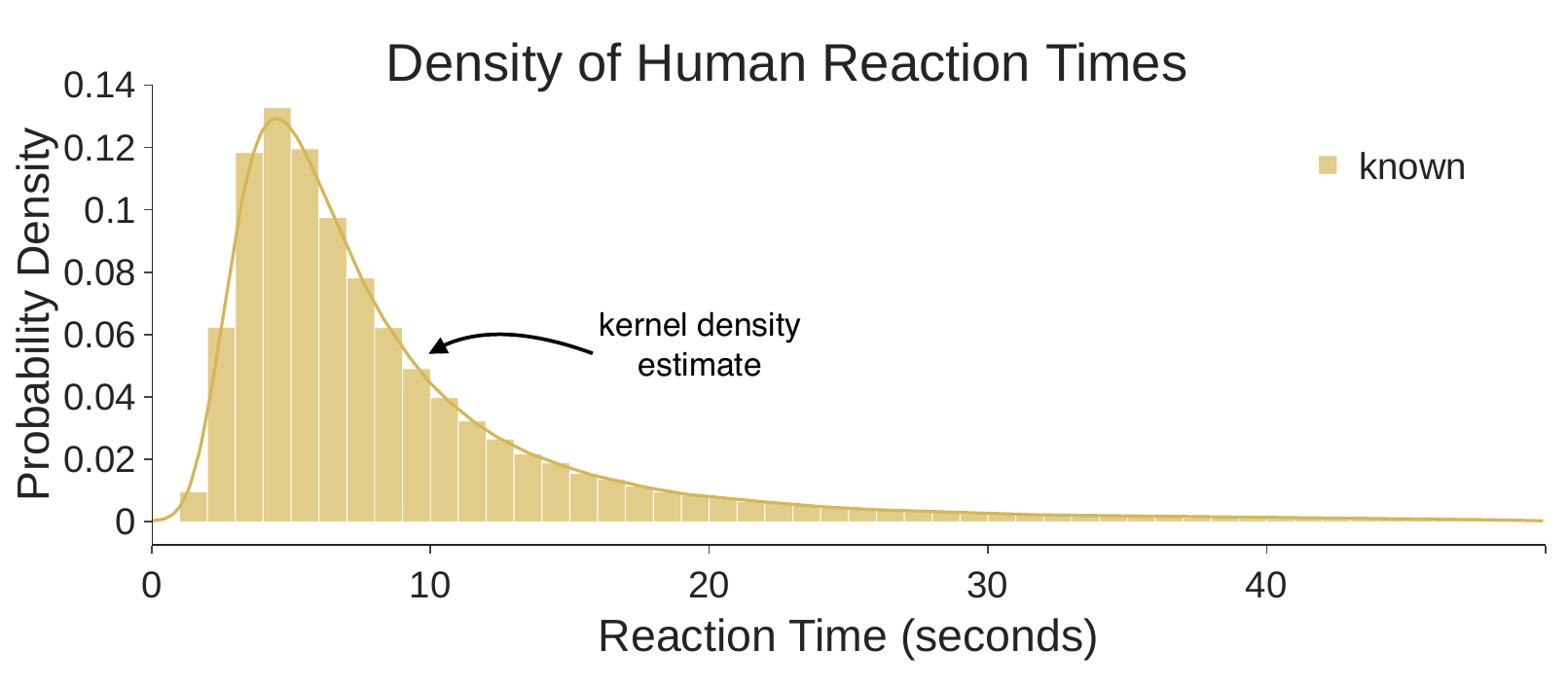}
    \caption{
        This histogram and kernel density estimate shows the distribution of human reaction time for the data collected for known samples.
        The X-axis shows the range of reaction times after thresholding the long tail, which removed outliers. The Y-axis shows the probability of occurrence.
        }
    \label{fig:rt_dist}
    \vspace{-5mm}
\end{figure}

\textbf{Summary of Collected Data.} We collected 211,074 RT measurements in total. After data cleaning and data pre-processing, we sampled 121,073 instance-level RTs for model training. We grouped the RTs that belong to the same image and calculated the average of them, and we used this average value to represent the RT for an image. Overall, there are 33,548 training samples from the 40 known classes, and 12,428 images among them have corresponding RTs.

After collecting the RT data, an important question for the subsequent machine learning work was whether to use class-level or sample-level behavioral information in the loss formulation. We started by looking at class-level RT by generating box plots summarizing the measurements by class pairings for each class (see Supp.~Mat.~Sec.~1.2). To be useful for a loss function, large relative differences between a class and its different pairings would need to be present. We observed that the variance in RT within a class across its various other-class pairings was small, and that the various pairings have similar minimum and median RT statistics. This indicates that there isn't enough RT information at the class-level to be useful for a loss function, as patterns of difficulty across different classes cannot be ascertained.

Next, we turned to the sample-level behavioral information. We plotted the distribution of RT measurements for these known samples, shown in Fig.~\ref{fig:rt_dist}.
Looking at the distribution, there is a large amount of variance across samples from all of the classes. This means that some samples have longer RT measurements while others have shorter ones, regardless of the classes they come from. Therefore the understanding and usage of RT has to be taken down to the sample-level for the design of the loss function.
Our RT data indicate that the human visual system is perceiving each sample differently. Using this finding as an intuition for machine learning, we make the following assumption: human RT is a significant indicator of the latent difficulty of perceiving an object relative to objects from other classes,  meaning these measurements will be useful or supervised training because the training set captures those relative comparisons.
\section{Training a Model with Variable Reaction Time}
\label{sec:approach}

\begin{figure}
    \centering
    \includegraphics[width=\columnwidth]{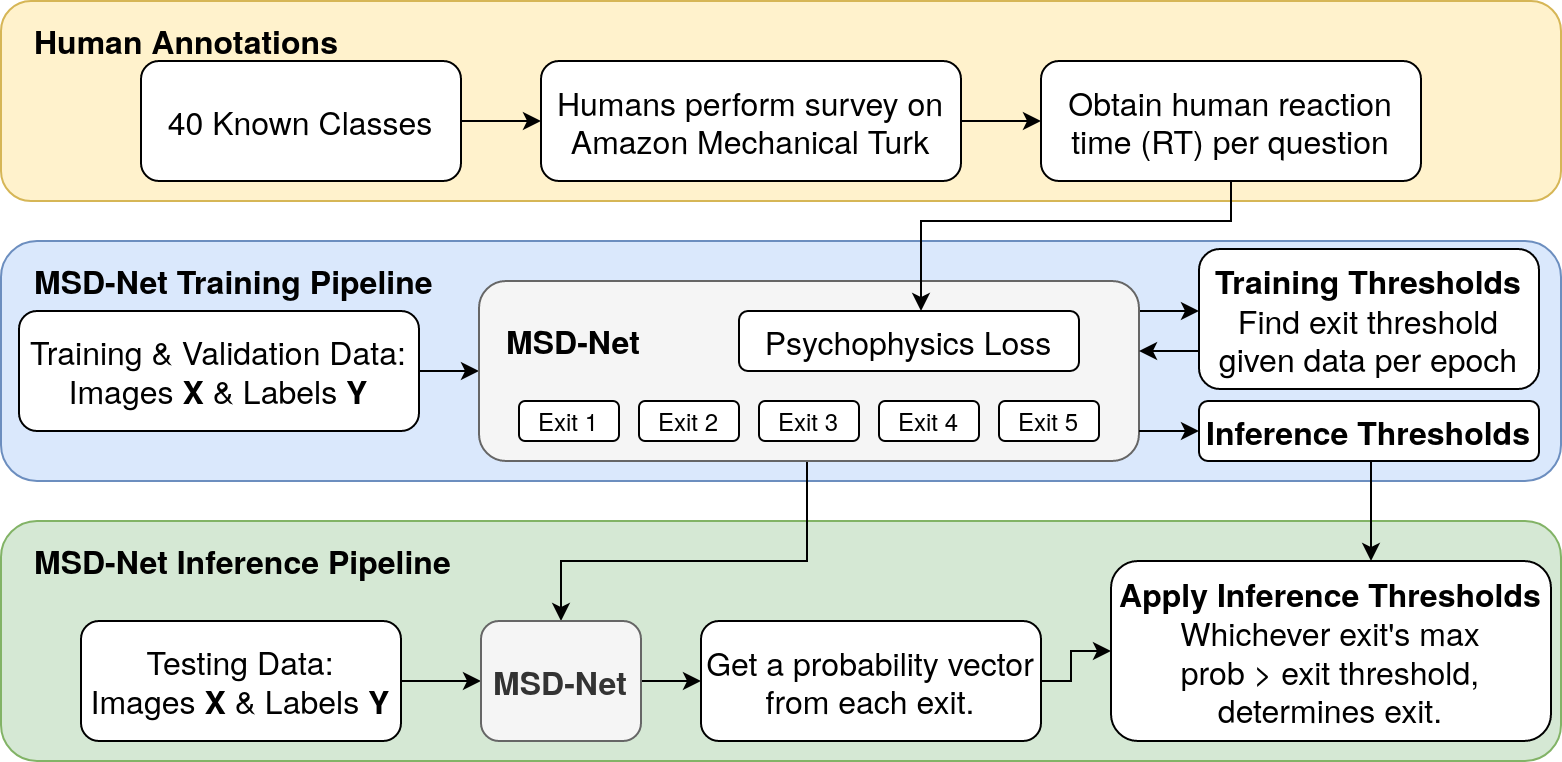}
    \caption{
        Pipeline depicting the three components of the proposed machine learning process.
        The yellow box represents the human study for collecting RT. 
        The blue box shows machine learning training. 
        The green box illustrates the testing process. See Supp.~Mat.~Sec.~3.1 for additional detail on these.
    }
    \label{fig:pipeline}
    \vspace{-5mm}
\end{figure}

With a large number of reaction time measurements associated with individual training points, we need some efficient way to make use of that data for training a deep network. The most straightforward approach has been to incorporate such data into a loss function~\cite{Scheirer_2014_TPAMI, grieggs2020measuring}.
Unexplored before in the OSR literature, however, is a tighter coupling between human and model behavior when it comes to the reaction time associated with inputs.
Thus we propose to enforce human behavioral fidelity in a model that supports dynamic reaction time during runtime (Fig.~\ref{fig:pipeline}). This is accomplished via a new psychophysical loss function that encourages the model to mimic per instance human RT.


\textbf{Multi-Scale DenseNet.}
Most network architectures use approximately the same amount of time to process all input images. Typically, networks only have one classifier at the end of the architecture, and it is extremely difficult, if not impossible, to discern any meaningful difference in model reaction time between samples.
To be able to support different model reaction times and to train the model with human RT data, we chose to use the Multi-Scale DenseNet (MSD-Net)~\cite{huang2018multiscale} architecture with a custom loss function. We also describe an experiment for a variant of ResNet-18 in Supp.~Mat.~Sec.~3.2.

Different from many other networks that only have one classifier at the end, MSD-Net has five classifiers in its architecture.
Utilizing these classifiers as model exits, we modify the network to output the predictions after each classifier, and use a thresholding strategy to determine whether the network is confident enough to decide which class a sample should be classified into, or whether a sample is unknown. 
We start training by setting all 5 thresholds corresponding to each exit to zero. To update these thresholds, we extract the prediction scores on validation data every 5 epochs at each exit, taking the median of all the prediction scores for a specific exit as its threshold. These updates are necessary because the model behavior changes over time, and probabilities become higher as the model better describes the data. 

We call these thresholds \emph{training thresholds}.
If the network produces a prediction score that is above the threshold and outputs a correct prediction, a sample is considered to be exiting the network. We consider it important that a sample has to satisfy both conditions during training because we need to enforce the constraint that the network confidently makes correct predictions. Training thresholds are updated based on the probability vectors from the exit loss  described below.

\textbf{Psychophysical Loss Formulation.}
The base loss used for training MSD-Net is the common multi-class classification cross-entropy loss.
The cross-entropy loss is the function
\begin{equation}
    \Loss_C(p, q) = -\sum_{y\in Y} p(y) \log q(y)
\end{equation}
 where $p$ is the labeled probability (one-hot vector) of the class $y$ from the closed set of classes $Y$ and $q$ is the MSD-Net predicted probability of the class $y$.
Here we add to the cross-entropy loss by incorporating RT measurements via two additional psychophysical loss functions.
This is accomplished via a weighted summation of the cross-entropy loss and a psychophysical loss.


The psychophysical loss is an exit loss
\begin{equation}
    \label{eq:three}
    \Loss_{E}(x, \hat{y}) = \left| {\Exit_{\mathrm{target}(x)}} - \Exit_{\hat{y}} \right| 
\end{equation}
where $\Exit_{\hat{y}}$ is the exit integer index of the MSD-Net prediction $\hat{y}$, and $\Exit_{\mathrm{target}(x)}$ is a lookup function that returns the target exit index for the sample $x$.
The exit loss is designed to push the MSD-Net's reaction time for the sample to be proportional to the sample's target exit based on a binning strategy informed by the human RT measurements.

To obtain the exit loss, the following two factors must be measured. \textbf{(1)} Expected exit: $\Exit_{\mathrm{target}(x)}$.
    This value is assigned by measuring a discrete distribution of human reaction time.
    After removing the outliers from the human RT measurements and obtaining the minimum and maximum values, the entire range is discretized into five bins 
    in accordance with the number of classifiers in the MSD-Net architecture.
    The cut-off thresholds defining the ranges covered by each bin are the quintiles of human RT in Fig.~\ref{fig:rt_dist} and are shown in Supp. Table 2.
    When a sample enters the network for training, the human RT associated with that sample is checked, and the corresponding exit index from the upper bound of the range is used as its $\Exit_{\mathrm{target}(x)}$.
    For instance, if a known sample has an average human RT of 5.5 seconds, its $\Exit_{\mathrm{target}(x)}$ will be 2.
    \textbf{(2)} Predicted exit for a sample determined by when it leaves the network: $\Exit_{\hat{y}}$.
    As a sample is processed through each classifier, the model checks the prediction scores as well as predicted label for that classifier.
    If the maximum prediction score produced is larger than the set threshold for an exit and the prediction is correct, the sample is considered to be leaving the network and the index of the exit is $\Exit_{\hat{y}}$.

The complete proposed psychophysical loss combines the  cross entropy loss and the exit loss as a weighted summation:
\begin{equation}
\begin{aligned}
    \Loss_{\Omega}(p, q, x, \hat{y}, \vec{\omega}) = &
        ~\omega_C \Loss_C(p, q) \\
        & +  \omega_E \Loss_{E}(x, \hat{y})
    \label{eq:loss}
\end{aligned}
\end{equation}
where $\vec{\omega} = [\omega_C, \omega_E] : \omega_i \in \mathbb{R}$ is a vector of weights that correspond to the hyperparameter weighting of each component.


\textbf{Training and Validation Pipeline.} We take an original training dataset of images that do not have associated RTs and split it into training and validation sets with a ratio of 70\% for training and 30\% for validation. Available samples that do have associated measured RT are proportionally split via the same ratio into training and validation sets and merged with the samples that do not have any associated RT. We utilize an existing implementation of MSD-Net (\url{https://github.com/kalviny/MSDNet-PyTorch}), and send training and validation data into the network to train it with the psychophysical loss. 
Optionally, the weights in $\vec{\omega}$ can be set via domain knowledge or hyperparameter optimization. In our experiments, all the weights were set to 1.0 to assess base performance.

\textbf{Testing Pipeline.} After obtaining a model from training, we perform post-processing for novelty detection. First, we need to determine the threshold for each exit. This leads to a different set of thresholds than the training thresholds; we call these \textit{inference thresholds}. To obtain this set of thresholds, the \textit{best model} needs to be identified first. During the training phase, we save a model as well as the training and validation accuracy associated with it at every epoch. After training is done, the model that produced the highest top-1 validation accuracy is selected as the best model. We then run validation data through this best model and calculate the median of all the predicted scores for each exit respectively; these median scores are considered to be the inference thresholds. They remain static from this point forward because the best model is finalized. We then run all test samples from both known and unknown classes through the selected model, saving the prediction scores from each exit. Lastly, the inference thresholds are applied to these scores to find out when the samples leave, with corresponding exit indices  recorded. 

The exit strategies are slightly different for known samples and unknown samples. For all the known samples, a sample correctly exits the network when the maximum score is larger than a given inference threshold, and when the network makes a correct prediction. If a sample does not exit from one of the first 4 exits due to being under threshold each time, it comes to the final exit, where there are three possibilities for it.
\textbf{(K1)} The maximum score of a sample is larger than the given threshold, and the prediction is correct: the sample exits and is correctly classified as a known class. \textbf{(K2)} The maximum score of a sample is larger than the given threshold, but the prediction is wrong: the sample exits and is classified as known, but associated with an incorrect class. \textbf{(K3)} The maximum score of a sample is smaller than the given threshold: the sample exits and is classified as unknown, making it a false negative.

As for testing unknown samples, there are only 2 possible cases for each sample. \textbf{(U1)} The maximum score is smaller than the given threshold at every exit: the sample is classified correctly as unknown. \textbf{(U2)} The maximum score at a particular exit is larger than the threshold given by that exit: the sample is wrongly classified as known, which makes it a false positive.

For testing known and unknown samples, we use the probability scores produced by the SoftMax function, and the exits solely decide whether a sample is known or unknown. During the testing phase, all samples are processed through the 5 exits and the probability scores from each exit are saved. The  scores are then post-processed to obtain the final classification results.



%
\section{Experiments and Results}

\textbf{Data and Evaluation Metrics.}
The training data breakdown can be found in Sec.~\ref{sec:study}.
For validation, 9,058 images are available from the 293 known classes. 
The test set consists of known and unknown images. The known test partition consists of the 293 classes seen in training providing a total of 336,453 new images.
The unknown test partition consists of 42 classes unseen at training --- all are considered to be part of one unknown class with 48,067 images. 

Although various OSR works~\cite{Neal_2018_ECCV, crosr, cacosr} use AUROC as a metric to evaluate  performance, we believe that it is somewhat flawed for this problem. 
Importantly, it obscures the performance achieved through the use of a single threshold by combining a large range of thresholds to produce one score. Moreover, it does not provide a recommendation for selecting the best threshold. AUROC can be used fairly as a metric to assess a model's performance during training or validation, but there should only be one threshold per classifier used in the testing phase, as would be done in operation. Knowing this, we chose to use fixed thresholds obtained via experiments with the validation data as our inference thresholds.

 
As OSR is a task built on top of novelty detection by adding multi-class classification for known samples, both the detection and classification components can and should be evaluated separately. Thus we include metrics for both when reporting our results. 
For novelty detection, we consider accuracy when testing unknown samples. For multi-class classification, we use accuracy score and consider top-$n$ accuracy for testing known samples. We report counts for True Positives (TP), True Negatives (TN), False Positives (FP), and False Negatives (FN), as well as F1 Score and the Matthews Correlation Coefficient (MCC) in the Supp.~Mat.~Sec.~3 for comparison.

\textbf{Results: Contribution of Each Loss Function.}
Since the proposed psychophysical loss is composed of an exit loss added on top of a cross-entropy loss, we trained two models separately: one model is trained just with cross-entropy loss and the other is trained according to Eq.~\ref{eq:loss} (the proposed loss). To compare the performance of our loss formulation with the previous published psychophysics loss \cite{grieggs2020measuring}, we also trained a model with this \textbf{performance loss}. It is noted as $\Loss_P$, and is added on top of cross-entropy loss as another baseline.

We train each of the models using a single GPU with a batch size of 16 images shaped 224x224 and an initial learning rate of 0.1 for 200 epochs. As for the optimizer, we use stochastic gradient descent (SGD) with a momentum of 0.9 and a weight decay of 0.0001. All other parameters are set the same as the original implementation, and we did not heavily tune these training parameters. Evaluation takes place in a 5-fold manner, training models with 5 different fixed seeds in all cases. The average accuracy over all folds is considered as the final result. 


\begin{table}
\centering
\resizebox{\columnwidth}{!}{
\begin{tabular*}{\columnwidth}{@{\extracolsep{\fill}}|c|c|c|c|c|}
\hline
 \textbf{Split} & \textbf{Accuracy} & $\Loss_C$ & $\Loss_C + \Loss_P$ & $\Loss_C + \Loss_E$ \\
\hline
  train & known top-1 & \textbf{72.56}$\pm0.47$ & \textbf{72.56} $\pm~0.27$ & 72.10$\pm0.42$ \\
  train & known top-3 & 85.01$\pm0.28$ & \textbf{85.09} $\pm~0.20$ & 84.35$\pm0.27$ \\
  train & known top-5 & 89.02$\pm0.22$ & \textbf{89.19} $\pm~0.15$ & 88.37$\pm0.22$ \\
\hline
  valid. & known top-1 & 53.98$\pm0.30$ & 52.90 $\pm~0.51$ & \textbf{60.00}$\pm0.27$ \\
  valid. & known top-3 & 68.76$\pm0.32$ & 68.36 $\pm~0.20$ & \textbf{72.12}$\pm0.28$ \\
  valid. & known top-5 & 74.67$\pm0.38$ & 74.18 $\pm~0.13$ & \textbf{77.34}$\pm0.18$ \\
\hline
  test & known  top-1 & 26.67$\pm7.76$ & 28.68 $\pm~6.91$ & \textbf{36.48}$\pm0.43$ \\
  test & known  top-3 & 36.40$\pm10.99$ & 38.90 $\pm~9.87$ & \textbf{46.70}$\pm0.73$ \\
  test & known  top-5 & 41.08$\pm12.24$ & 43.76 $\pm~11.14$ & \textbf{51.00}$\pm0.89$ \\
  test & unk. top-1 & 21.05$\pm0.63$ & 23.95 $\pm~2.98$ & \textbf{54.23}$\pm0.48$ \\
\hline
\end{tabular*}}
    
    \vspace*{1mm}
    \caption{
        Results for three variations of the loss used to train MSD-Net; scores are accuracy (\%).
        For each loss, we run experiments 5 times using 5 different seeds for training; the scores shown for each experiment are the average of 5 runs with standard error. $\Loss_P$ is a performance loss~\cite{grieggs2020measuring} described in Sec.~2 of the Supp.~Mat. 
        Additional results, including results for a model trained with all 3 losses and complete results for all 5 runs, are in Sec.~3 of the Supp.~Mat.
    }
\label{tab:three_baselines}
\vspace{-8mm}
\end{table}

Table~\ref{tab:three_baselines} shows the training, validation and testing results for this experiment.
All three configurations of the loss demonstrate roughly the same accuracy on known samples during training. When performance loss is added to cross-entropy loss, there is a noticeable uptick in test accuracy for both known and unknown samples. This indicates that measured RT is helping the trained model to generalize in the OSR setting. The increased performance in this case on known samples is consistent with prior work~\cite{grieggs2020measuring}, but the increase in performance on unknown samples is a new finding. Better conditioning of the MSD-Net exit behavior yields even more performance improvement. 
For top-1 validation accuracy, the proposed loss 
of Eq.~\ref{eq:loss} outperforms the cross-entropy loss by 6.02\%, and the combined cross-entropy and performance loss by 7.1\%. Even more significant improvement is demonstrated in the testing phase.
When testing on known classes, the proposed loss results in a top-1 accuracy of 36.48\%,  outperforming the two other loss configurations (26.67\% and 28.68\%). When detecting unknown samples, the proposed loss largely outperforms the others with 54.23\% accuracy, compared to 21.05\% and 23.95\%. These results show that our proposed loss leads to models that are able to maintain good performance when testing with multiple known classes while simultaneously rejecting unknown samples.

\textbf{Results: Other Baselines.}
We also compared our method with eight other OSR methods from the literature. With respect to standalone classifiers, we evaluated SVM~\cite{svm} with thresholding, $P_I$-SVM~\cite{pi-svm}, W-SVM~\cite{wsvm}, and the EVM~\cite{Rudd_2018} with  MSD-Net features. With respect to deep learning-based methods, we evaluated OpenMax~\cite{Bendale_2016_CVPR}, OSRCI~\cite{Neal_2018_ECCV}, CROSR~\cite{crosr}, and CAC-OSR~\cite{cacosr}. See Supp.~Mat.~Sec.~2 for descriptions of these approaches. For these experiments, we used the same dataset used to evaluate MSD-Net training with different loss configurations. As we did not heavily tune hyperparameters in our proposed approach to avoid overfitting, for a fair comparison we also refrained from doing so for all baseline methods. 
Table~\ref{tab:all_results} shows results on known samples and unknown samples from the test set for all methods, including the different configurations of MSD-Net training. 





While most of the methods only achieve performance that is only a little better than making a random choice among the 293 known classes, the different configurations of MSD-Net training achieve excellent top-1 accuracy when testing on known samples, with the proposed loss yielding the best result.
The EVM has the highest accuracy when tested with unknown samples, but when we look at the accuracy for known samples, we notice that it tends to classify everything as unknown.
Similar to the EVM, CROSR and CAC-OSR also classify most of the samples as unknown, leading to a high accuracy when considering just the unknowns, but very poor accuracy for known samples.
Reversing this trend, OSRCI classifies most of the classes as known, which indicates it produces high prediction scores for both known and unknown samples, but is not able to distinguish between known and unknown samples, and at the same time classifying known samples into wrong classes. SVM and OpenMax are more balanced when it comes to OSR and are reasonably accurate in detecting unknown samples, but like other algorithms, they failed to perform well when assigning known samples to the correct class. Only the MSD-Net configurations are able to maintain good performance for both novelty detection and classification. 

\begin{table}[t]
  \centering
  \begin{tabular*}{\columnwidth}{@{\extracolsep{\fill}}|c|c|c|}
  \hline
    \textbf{Algorithm} & 
    \textbf{Test Unknown Acc.} & 
    \textbf{\begin{tabular}{@{}c@{}} Test Known \\ Top-1 Acc\end{tabular}} \\
    \hline
    SVM~\cite{svm} & 55.04\% $\pm~0.76$ & 0.60\% $\pm~0.07$ \\
    \hline
    $P_I$-SVM~\cite{pi-svm} & 8.31\% $\pm$ 0.12 & 0.28\% $\pm$ 0.02 \\
    \hline
    W-SVM~\cite{wsvm} & 9.21\% $\pm$ 0.15 & 0.09\% $\pm$ 0.01 \\
    \hline
    EVM~\cite{Rudd_2018} & \textbf{{\color{red}{99.39}\%}} $\pm$ 0.00  & 0.45\% $\pm$ 0.00 \\
    \hline
    OpenMax~\cite{Bendale_2016_CVPR} & 32.41\% $\pm$ 0.04 & 0.39\% $\pm$ 0.00\\
    \hline
    OSRCI~\cite{Neal_2018_ECCV} & 1.11\% $\pm$ 0.01 & 0.36\% $\pm$ 0.00 \\
    \hline
    CROSR~\cite{crosr} & 85.78\% $\pm$ 0.00 & 1.92\% $\pm$ 0.00 \\
    \hline
    CAC-OSR~\cite{cacosr} & 94.55\% $\pm$ 0.00 & 0.05\% $\pm$ 0.00 \\
    \hline
    MSD-Net $\Loss_C$ & 21.05\% $\pm$ 0.63 & 26.67\% $\pm$7.76 \\
    \hline
    MSD-Net $\Loss_C + \Loss_P$~\cite{grieggs2020measuring} & 23.95\% $\pm$ 2.98 & 28.68\% $\pm$ 6.91 \\
    \hline
    MSD-Net $\Loss_C + \Loss_E$ (Eq.~\ref{eq:loss}) & 54.23\% $\pm$ 0.48 & \textbf{36.48\%} $\pm$ 0.43 \\
    \hline

  \end{tabular*}
  
  \vspace*{1mm}
  \caption{Testing results for the proposed loss and other baselines. Bold numbers indicate best performance for a metric. Results in red highlight spurious best results due to a classifier biased towards determining samples are novel. }
  \label{tab:all_results}
  \vspace{-8mm}
\end{table}

The poor performance of the baseline approaches on the ImageNet335 OSR task was surprising. The most significant cause is likely dataset size. As emphasized earlier, OSR research tends to use a handful of small datasets with far fewer than 335 classes. Testing performance is known to drop when the number of classes increases~\cite{geng2020recent}. Further, we have intentionally curated the dataset used in this paper to provide relatively few examples per class in order to assess the impact of adding behavioral data. Larger-scale OSR work has made use of Tiny ImageNet~\cite{Neal_2018_ECCV, crosr,cacosr} or full ImageNet pre-training~\cite{Bendale_2016_CVPR,Rudd_2018} with access to hundreds of thousands of labeled images for training --- an order of magnitude more data than what is available in our experiments for training. Our proposed approach gains information from the behavioral data that partially compensates for having fewer training samples per class, which otherwise heavily degrades performance. 

Secondary reasons for poor performance include input image size sensitivity for deep learning-based models and feature size constraints for standalone classifiers. In the former case, to be able to use this data with the baselines, we had to downscale the images to match the input size required by each baseline method, which potentially led to a loss of information when they were trained. In the latter case, we had to use PCA to reduce the dimensionality of the MSD-Net features for a few of the baseline methods (details can be found in Supp.~Mat.~Sec.~2), including: SVM, W-SVM, $P_I$-SVM and EVM. Such brittleness needs further attention in OSR work. 


\section{Discussion and Future Work}


While our method is applied to a limited number of training samples in this study, it is also possible to utilize the psychophysical loss function on larger datasets (including those with outlier samples~\cite{Deep_Anomaly_Detection_with_Outlier_Exposure}). We argue that human RT can still improve model performance, because it provides extra information for deciding whether a sample should be known or unknown. It is worth mentioning that with a larger training dataset, a larger number of RT measurements may be necessary. Limitations for the proposed method are focused around the data collection. For instance, if a large amount of human data is required then the data collection process can be very time consuming. Further, the collected human data is task specific, and is difficult to transfer to other applications. However, these are opportunities for future work.

Turning to humans as a reference point for OSR problems is a promising new direction of work, and our results hint at the potential other experimental practices from the field of psychology might have in providing information beyond simple labels to supervised learning. 
Gaze measurement has already been suggested as one data stream~\cite{zhang2018agil}. 
More intriguing though is the measurement of pupil dynamics. A recent study has shown that movement of the pupil can be used as an index of mental effort exertion~\cite{pupil}, and thus could be another indicator of the patterns of error in training data.

There is also work that looks at the relationship between human behavior and machine learning models beyond RT, suggesting that human perception does not always align with model decisions. Geirhos et al.~\cite{Generalisation_in_humans} suggest that humans and computer vision models process information in different ways, and Kotseruba et al.~\cite{saliency_model} argue that training CNNs with psychophysical data does not improve the performance of saliency models. 
This research provides a foundation for exploring what types of tasks can benefit from psychophysics and alternative ways to incorporate human behavior into  algorithms. With RT, we are just scratching the surface.

\ifCLASSOPTIONcompsoc
  \section*{Acknowledgments}
\else
  \section*{Acknowledgment}
\fi

This research was sponsored in part by the National Science Foundation (NSF) grant CAREER-1942151 and by the Defense Advanced Research Projects Agency (DARPA) and the Army Research Office (ARO) under multiple contracts/agreements including HR001120C0055, W911NF-20-2-0005, W911NF-20-2-0004, HQ0034-19-D-0001, W911NF2020009. The views contained in this document are those of the authors and should not be interpreted as representing the official policies, either expressed or implied, of the DARPA or ARO, or the U.S. Government.

\ifCLASSOPTIONcaptionsoff
  \newpage
\fi



\bibliographystyle{IEEEtran}
\bibliography{egbib.bib}
\end{document}